%% file: main.tex
\documentclass{article}

\usepackage{PRIMEarxiv}

\usepackage[utf8]{inputenc} 
\usepackage[T1]{fontenc}    
\usepackage{hyperref}       
\usepackage{url}            
\usepackage{booktabs}       
\usepackage{amsfonts}       
\usepackage{nicefrac}       
\usepackage{microtype}      
\usepackage{lipsum}
\usepackage{fancyhdr}       
\usepackage{graphicx}       
\graphicspath{{media/}}     
\usepackage{float}
\usepackage{amsmath}
\usepackage{amssymb}
\usepackage{multirow}
\usepackage{multicol}
\usepackage{makecell}
\usepackage{boldline}
\usepackage{subfig}
\usepackage{authblk}

\usepackage{xcolor, soul}
\sethlcolor{yellow}

\input{01-info}

\begin{document}

\maketitle    

\input{02-abstract}

\section{Introduction}
\label{sec:intro}

\input{03-intro}

\section{Technical Approach}
\label{sec:methods}
\input{04-methods}


\section{Results and Discussion}
\label{sec:results}
\input{05-results}


\section{Limitations and Future Work}
\label{sec:limitations}
\input{06-limitations}


\section{Conclusion}
\label{sec:conclusion}
\input{07-conclusion}


\section{Acknowledgment}
This research was funded by Air Force Research Laboratory S111068002.

\input{main.bbl}
\appendix
\label{sec:appendix}

\input{08-appendix}
\end{document}

%% file: 01-info.tex
\title{Scalar Field Prediction on Meshes Using Interpolated Multi-Resolution Convolutional Neural Networks}

\author[1]{Kevin Ferguson}
\author[2]{Andrew Gillman}
\author[2]{James Hardin}
\author[1]{Levent Burak Kara}

\affil[1]{Carnegie Mellon University\\
  5000 Forbes Avenue\\
  Pittsburgh, PA 15213\thanks{Address all correspondence to lkara@cmu.edu}}
\affil[2]{Air Force Research Laboratory\\
  2977 Hobson Way\\
  Wright-Patterson Air Force Base, OH 45433}

%% file: 02-abstract.tex
\begin{abstract}
Scalar fields, such as stress or temperature fields, are often calculated in shape optimization and design problems in engineering. For complex problems where shapes have varying topology and cannot be parametrized, data-driven scalar field prediction can be faster than traditional finite element methods. However, current data-driven techniques to predict scalar fields are limited to a fixed grid domain, instead of arbitrary mesh structures. In this work, we propose a method to predict scalar fields on arbitrary meshes. It uses a convolutional neural network whose feature maps at multiple resolutions are interpolated to node positions before being fed into a multilayer perceptron to predict solutions to partial differential equations at mesh nodes. The model is trained on finite element von Mises stress fields, and once trained it can estimate stress values at each node on any input mesh. Two shape datasets are investigated, and the model has strong performance on both, with a median R-squared value of 0.91. We also demonstrate the model on a temperature field in a heat conduction problem, where its predictions have a median R-squared value of 0.99. Our method provides a potential flexible alternative to finite element analysis in engineering design contexts. Code and datasets are available at: \url{https://github.com/kevinferg/sfp-cnn}.
\end{abstract}

%% file: 03-intro.tex
Geometry-based optimization and design problems are prevalent in engineering. Engineers frequently must design the shape of a part and then perform an analysis on the part to determine what aspects of the design should be modified. This iterative process is well-established \cite{cad_fea}, but it is often inhibited by the slow speed of the analysis stage, which requires setting up a finite element simulation; assigning all of the necessary material properties, boundary conditions, loads, and other parameters; and waiting for a finite element software to mesh the geometry, assemble stiffness/load matrices, and solve for the requested values. Techniques to streamline this process are often sought-after in engineering design settings \cite{design_optimization,design_process}. 

One tool that would expedite the design process is a fast way to locate weaknesses in candidate designs, as this would let an engineer save expensive computer analyses for a more finalized design. Data-driven scalar field prediction methods solve this problem by predicting desired quantities at every point in a particular domain. In the case of shape design, for example, this may involve predicting a failure probability or equivalent stress at every node in the part's mesh. The task is not trivial, as small geometric changes can result in drastic field changes. However, by training across many examples, data-driven methods can learn physical phenomena, forming surrogate models that can predict finite element results \cite{truss1,magnet,stressnet,stressgan,sla_stress,truss2}.

Nie et al. \cite{stressnet} and Jiang et al. \cite{stressgan} propose deep-learning methods that solve for stress fields under varying geometries, boundary conditions, and loads. However, these approaches are image-based, so all inputs, including geometry, are encoded as binary image representations. The fidelity of the resulting finite element computations are highly restricted by this approach. Any fields computed for these shapes would be fixed to a single grid of pixels, significantly limiting the application potential of these methods, since most often shapes in engineering contexts are not designed with these limitations. What these approaches lack is flexibility in the representation of their shapes, which would be afforded by adopting a point cloud-, graph-, or interpolation-based method.

Qi et al. \cite{pointnet,pointnet2} present methods for point cloud classification and segmentation. Point cloud segmentation is similar to scalar field prediction, and the method has been used by Kashefi et al. \cite{fluid_flow} for the same purpose. Mesh segmentation techniques have shown good results on segmenting 3-D datasets \cite{mesh_seg,mesh_label}, providing evidence that predicting scalar fields on structures should be a feasible task.

Another popular approach to solve similar problems is to use a graph neural network (GNN) to make predictions on nodes of a graph \cite{gcn,dynamic}. These often make use of ``message-passing'' by iteratively aggregating graph neighborhood information at each node along graph edges, to learn representations of local graph structure -- these are then used for predicting node embeddings (such as scalar fields). GNNs for physics predictions are used more often for computing updates of dynamic simulations than for predicting static scalar values \cite{gnn_physics}. More recently, Meyer et al. \cite{direct} perform direct-time GNN predictions for Computational Fluid Dynamics (CFD), which is more comparable to the scalar field prediction we wish to investigate. In the domain of solid mechanics, Whalen and Mueller \cite{truss2} studied GNN-based surrogate models for predicting truss displacements, while Maurizi et al. \cite{microstructure} use GNNs to predict stress, strain, and deformation fields for different meta-materials. GNNs are powerful in that input graph topology can be arbitrary -- however, with a large graph or fine mesh, the depth of these networks needs to be very high, leading to substantial computational expenses \cite{deep,mesh}. Furthermore, GNNs tend to over-smooth solutions, and therefore may be less apt for predicting fields with steep spatial gradients \cite{smooth}. 

Some other data-driven scalar field prediction methods combine finite element methods with machine learning methods \cite{fea,cfd}, but these models requires performing a modified finite element simulation as part of the prediction, when a pure surrogate model is preferred.


Several works have explored combining features at multiple resolutions, especially using interpolation, in deep learning methods. Applications of such methods range from engineering to computer graphics. M{\"u}ller et al. \cite{hash} use trainable tables of features at multiple resolutions to represent multiple fields in 2-D and 3-D in computer graphics. Pandey and Ramakrishnan \cite{superres} perform image super-resolution by incorporating interpolation into a CNN. Such approaches contain similar mechanisms to the one found in our proposed method.

In this work, a model that predicts the Von Mises stress field in a static structural problem for an arbitrary mesh by training on the results of many finite element simulations is described. We propose a method that takes in a signed distance field image representation of a part's geometry, and processes this image with a convolutional neural network (CNN). The CNN has interpolation built-in, and by repeatedly downsampling and interpolating, it constructs feature vectors at each node that are passed through a multilayer perceptron (MLP) to predict stress at each node.

We demonstrate the model's performance on two large shape datasets. The $R^2$ values of predictions for meshes in the training set, as well as for previously unseen meshes, are shown and the predicted fields are visualized. We also investigate the interplay between network depth and prediction quality. We then validate the method on a heat transfer problem and investigate possible weaknesses in the model. 

The paper's main contributions are:
\begin{enumerate}
    \item A public dataset of 2-D shapes with complex internal geometries, which cannot be easily defined by a small set of parameters
    \item A convolutional neural network architecture that incorporates multi-resolution interpolation to make predictions at arbitrary locations
    \item A surrogate model for structural analysis problems that takes in a signed distance field and predicts von Mises stress throughout a part
\end{enumerate}

%% file: 04-methods.tex
\begin{figure}
\centering
\includegraphics[width=.6\textwidth]{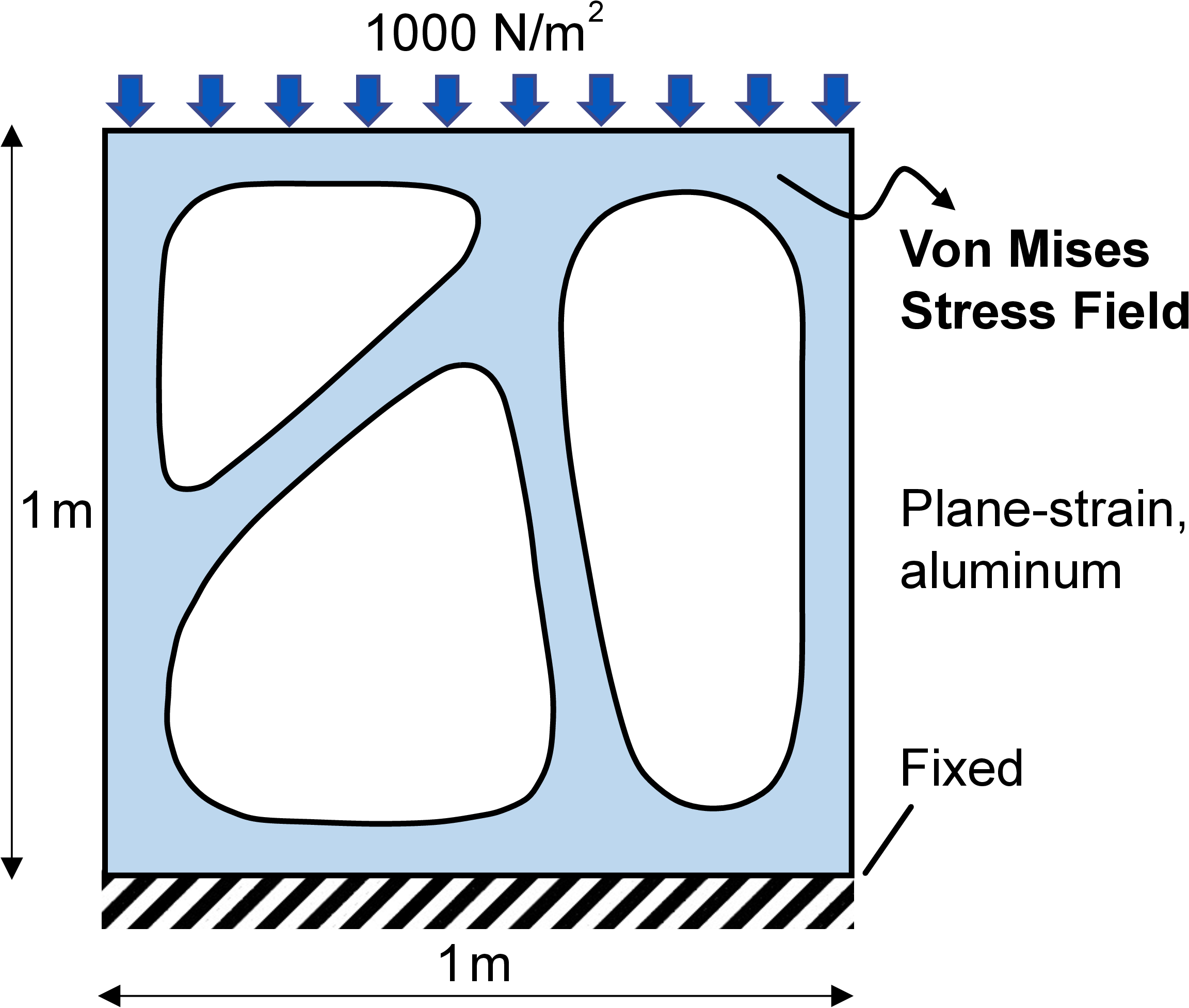}
\caption{The 2-D problem of interest: compression of aluminum part with variable internal geometry}
\label{fig:2d_problem} 
\end{figure}

\subsection{Problem}

As a test problem, we investigate a 2-D plane-strain scenario where a distributed load is applied to the top of an aluminum square domain that has a fixed bottom boundary, in which the goal is to predict von Mises stress. The details and dimensions of this problem are depicted in Fig.\ref{fig:2d_problem}. The material and dimensions chosen are arbitrary; we assume that predicting the von Mises stress field for this 2-D problem will yield results of similar quality to other physical fields without loss of generality. Stress fields exhibit steeper spatial gradients than, for example, temperature fields, and should therefore be more challenging to predict; to bolster this claim, we will also show our model's performance on a heat transfer problem in the Results section. Additionally, the  von Mises stress field has immediate practical relevance to engineering design problems, as comparing von Mises stress with material strength is commonly used to determine whether yielding will occur \cite{vmstress}. Note that the geometry on the interior of the square domain varies, as predictions will be made on different meshes.

\subsection{Datasets}
\label{sec:datasets}

\begin{figure}
    \centering
    \includegraphics[trim = 0in 0in 0in 0in, clip,width=.9\textwidth]{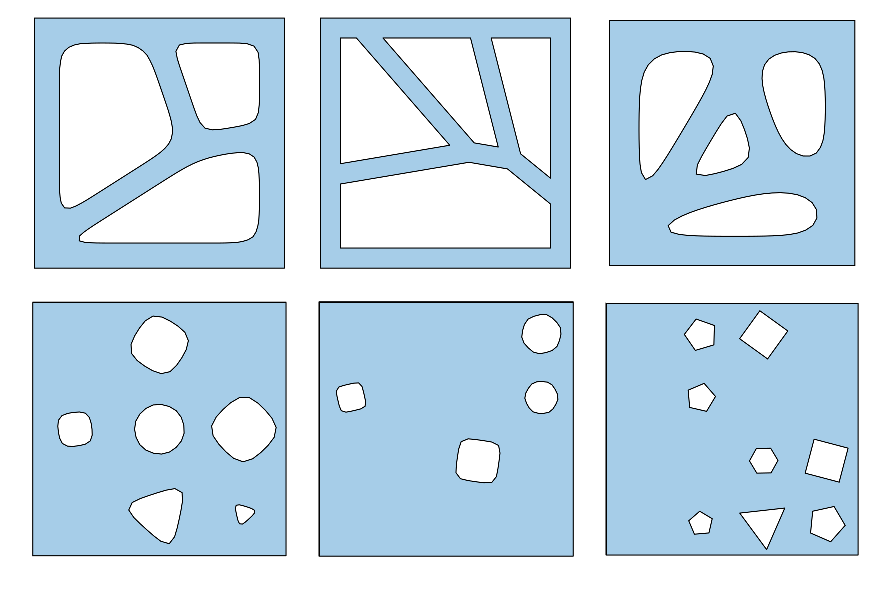}
    \caption{Examples of shapes in the Voronoi Set (top) and Lattice Set (bottom)}
    \label{fig:examples}
\end{figure}

Scalar field prediction models were trained for two 2-D shape datasets, which will be referred to as the \textit{Voronoi Set} and the \textit{Lattice Set}. Figure \ref{fig:examples} depicts a example shapes from each. The datasets were designed such that interior geometries would have varying topology and drastically different qualitative features. In an engineering design setting, freedom to laterally explore large design spaces like those in these datasets is desirable. For both datasets, a small set of parameters is not sufficient to fully describe the variation across meshes. Furthermore, the scalar fields that can exist on these complex geometries are not trivial to predict, making these datasets a good test-bed for examining the strength of the proposed features. There is not a one-to-one correspondence between any pair of meshes in either dataset -- the graph structures and individual node locations can be fully arbitrary. 

Each dataset is partitioned into training and testing sets with approximately 80\% of shapes used for training. Another ``out-of-distribution" set is generated for each, which are generated using parameters outside the range seen in training. This will be treated as a separate testing dataset; evaluating each model on these out-of-distribution shapes will help determine how significantly the proposed method is able to extrapolate beyond the geometries seen during training.

The results highlighted in Sec.\ref{sec:results} will focus mostly on a model trained with a combination of these datasets, referred to as the \textit{Combined Set}. For the Combined Set, each partition of the Voronoi Set has been merged with its respective partition of the Lattice Set. Finite element solutions were computed using MATLAB's PDE toolbox \cite{matlabpde} to solve for von Mises stress for the posed problem on each geometry. Meshes contain linear triangular elements with a maximum edge length of 0.025m.

\subsubsection{Voronoi Set}

\begin{figure}
    \centering
    \includegraphics[width=0.8\textwidth]{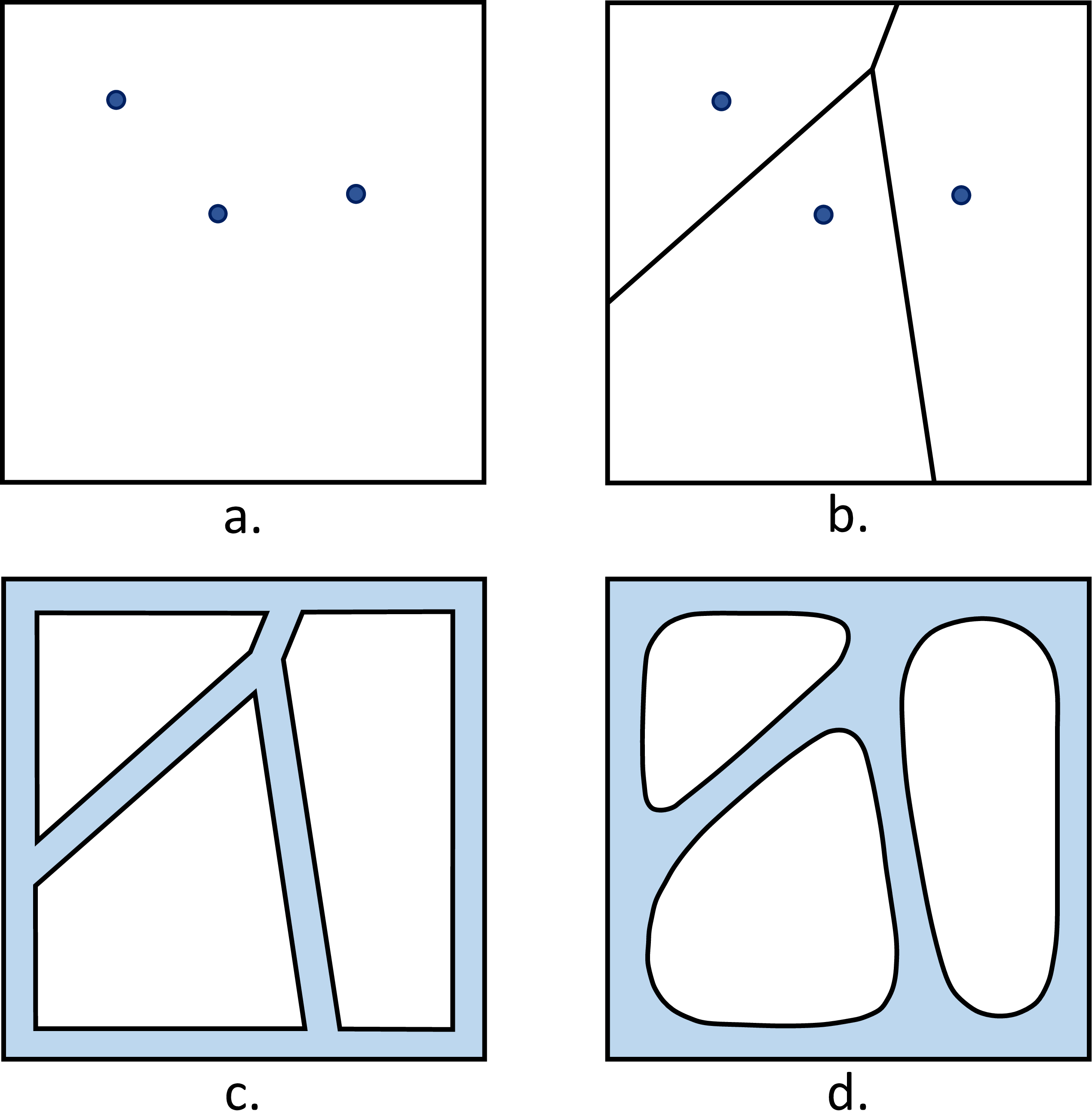}
    \caption{Generating the Voronoi Set: a. Random point selection; b. Voronoi tessellation; c. Pore creation; d. Smoothing}
    \label{fig:voronoi_set}
\end{figure}

Kou and Tan \cite{geometry} describe a method for generating porous structures by computing B-spline curves whose control points are vertices of Voronoi cells in a 2-D domain, and then randomly merging these curves to become pores. The Voronoi Set is inspired by this method; first, a set of points are randomly sampled from a square domain and Voronoi cell boundaries are computed. A buffer around each cell boundary is generated in order to control the wall thickness between the pores, after which Laplacian smoothing \cite{smoothing} can be performed to round the corners. The method is illustrated in Fig.\ref{fig:voronoi_set}. In addition to the random point coordinates, three parameters control this dataset: the number of holes, wall thickness, and degree of smoothing. In the Voronoi Set, the parameter ranges are: 3-4 holes, wall thickness from 0.10-0.18, and smoothing degree from 1-20. In the out-of-distribution Voronoi Set, wall thickness parameters are as low as 0.06 or as high as 0.22. There are 800 training, 200 testing, and 200 out-of-distribution Voronoi Set shapes.

\subsubsection{Lattice Set}

In the second dataset, the Lattice Set, each shape is built as follows: First, a square lattice of points on the domain is selected. Of these points, a random subset of them are selected to be the center locations of holes. Each hole takes the form of a smoothed polygon with a random orientation, smoothness, size, and side count. Within the training set, the lattice is either $3\times3$ or $4\times4$, the number of holes is between 4 and 8, inclusive; polygon side count ranges from 3 to 6; and polygon smoothness degree ranges from 1 to 15. For out-of-distribution Lattice Set shapes, hole counts are as low as 2 or as high as 10. There are 800 training, 200 testing, and 160 out-of-distribution Lattice Set shapes.

\subsection{Model}

\begin{figure}[t]
    \centering
    \includegraphics[width=\textwidth]{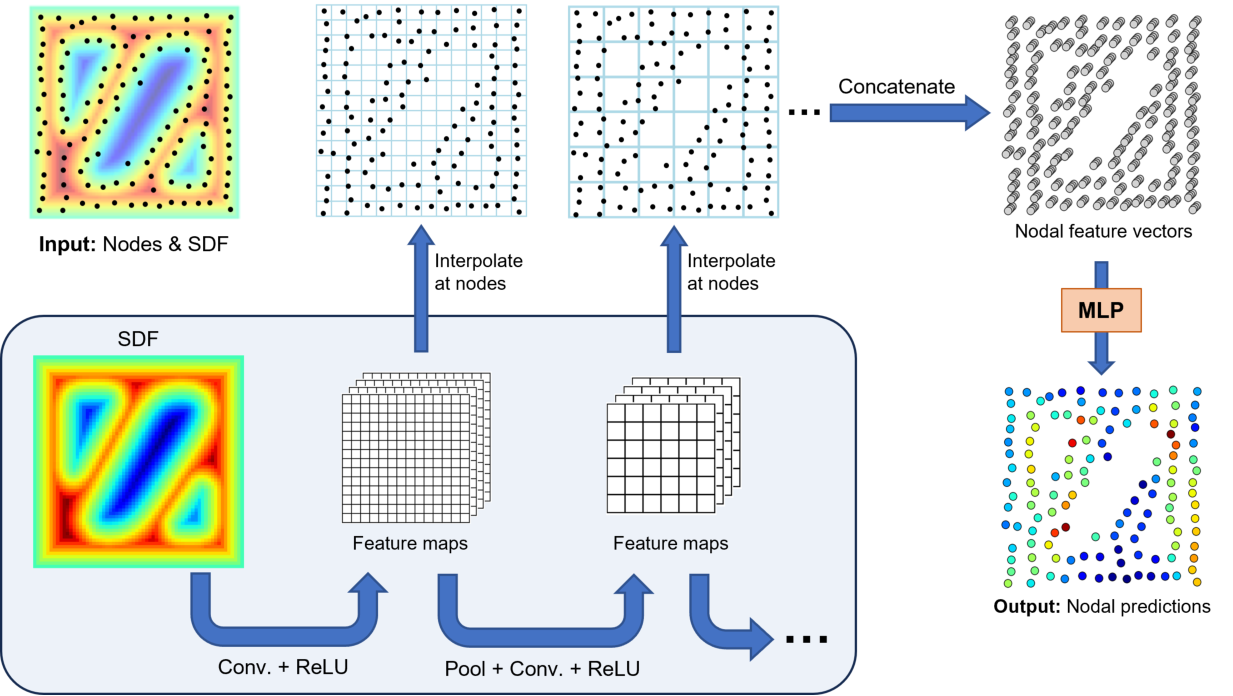}
    \caption{The proposed model, a neural network that performs several steps of convolution, pooling, and interpolation to construct multi-resolution feature vectors at each node, before using an MLP to predict the nodal scalar values.}
    \label{fig:model_specific}
\end{figure}

Our model, shown in Fig.\ref{fig:model_specific} is a convolutional neural network that takes as input both an image representing the shape geometry and the nodal coordinates of the shape's mesh. The output is a scalar value at each node, the collection of which we refer to as a scalar field. 

Just as each node can be mapped to a particular location on the geometry image, any other image can be overlaid on the mesh; the value of the image at any node can be computed by interpolating the values of nearby pixels, e.g. with bilinear interpolation. This mechanism is employed by our model to give predictions at each node, instead of only on an image-like grid of fixed resolution. Because the von Mises stress field we seek to predict is known only at the coordinates of each node, outputting node-wise values in this way is necessary.

First, the input image is fed into a 2D convolution, which outputs an array of feature maps the same size as the original image. ReLU activation is applied to the feature map outputs of each convolutional layer. These feature maps are then treated as images and interpolated at mesh nodes to give the nodal features at this resolution. More information on this interpolation is provided in Section \ref{sec:interp}.

Next, the feature maps are reduced in size with max-pooling, before applying another 2D convolution followed by ReLU. These feature maps, although smaller, can still be mapped onto the mesh space. Therefore, they are also interpolated to return additional nodal features corresponding to a coarser level of detail. 

This process is repeated multiple times, resulting in several feature vectors at each node, corresponding to several different resolution levels. Each of these feature vectors are concatenated resulting in a single large vector of features at every node. By combining features at several resolutions, this stage of the network has the same residual effect as skip-connections found in similar segmentation networks like U-Net \cite{unet}. That is, it preserves detail at higher resolutions, but still contains more global low-resolution information.

Finally, the concatenated features are passed node-wise through an MLP to give the output at each node. This MLP serves the same purpose as the final MLP stage of large point cloud segmentation networks \cite{pointnet}; it does the final processing on node-wise feature vectors to output a final prediction at each node.

\subsubsection{Signed Distance Field}
\label{sec:sdf}

Let $\Omega$ represent the interior domain of a shape, while $\Omega^-$ represents the shape's exterior. Let $\partial\Omega$ be the part's boundary -- that is, both the outer square and the edges of each pore for our datasets. The Signed Distance Field (SDF) at an arbitrary point $\boldsymbol{x}$ is given by:

\begin{equation}
    \text{SDF}(\boldsymbol{x}) = 
    \begin{cases}
    d(\boldsymbol{x}, \partial\Omega) & \boldsymbol{x} \in \Omega \\
    -d(\boldsymbol{x}, \partial\Omega) & \boldsymbol{x} \in \Omega^-,
    \end{cases}
\end{equation}

where $d(\boldsymbol{x}, \partial\Omega)$ is the Euclidean distance from the point $\boldsymbol{x}$ to the nearest point on the boundary $\partial\Omega$ \cite{sdf}. The SDF is positive for points on the interior of the geometry, negative for exterior/void regions, and 0 on the boundary. On an arbitrary domain, the SDF can said to represent a solution to a special case of the Eikonal equation \cite{eikonal}, which can be computed efficiently through the fast marching method \cite{fastmarching}. This ``distance to a boundary'' field is a useful value for representing geometry locally, and as such, similar fields are frequently used in image processing applications \cite{distance}. This field contains more information at any point than a simple binary representation of the domain. Hence, the SDF sampled on a grid of size $64\times64$ within the unit square is used as a means of providing the shape of each part as input to the model.

\subsubsection{Interpolation}
\label{sec:interp}
The primary method by which our model is able to make predictions at arbitrary node locations despite the underlying approach using image convolution is interpolation from an image space to the node space. For every shape in the dataset, the SDF has been sampled on a uniform $64\times64$ grid with corners at the corners of the unit square. However, since nodes may be located anywhere within the unit square, interpolation is necessary. For this, we look to bilinear interpolation, which determines, for a given node, which four pixels on the image are nearest it, and applies linear interpolation in the x- and y-directions to estimate what value the image would have at the location of the node.

For each interpolation step, we assume that the bottom left, bottom right, top left, and top right entries of each feature map have $(x,y)$ coordinates of $(0,0)$, $(1,0)$, $(0,1)$, and $(1,1)$ respectively. All other feature map entries can then be considered the vertices of a uniform grid of 2-D cells making up the unit square. Then any node on the target shape can be mapped to a location within one of these cells. The node's feature vector is calculated by performing bilinear interpolation using the cell's corner vertex feature map values. We have implemented a function in PyTorch \cite{pytorch} to efficiently perform this interpolation for multiple feature maps to a large batch of nodes concurrently.

For 3-D problems, an analogous approach enables interpolation using the 8 vertex values of the box in which a query point lies, as in \cite{hash}. Smoother schemes such as Hermite interpolation and similar techniques based on ``smoothstep" interpolation, can also be applied, which will lead to smoother field results but may take longer to compute \cite{interp}.

\subsection{Training Details}

The network takes in a $64\times64$ input SDF image. This is passed through $5\times5$ convolution with stride 1 and zero-padding 2, and a ReLU activation is applied. The resulting feature map contains 20 channels, which are interpolated to each node, giving the first 20 nodal features. Next, the 20-channel feature maps go through a $2\times2$ max-pooling operation with stride 2 to result in a $32\times32$ image. Additional convolution, ReLU, and interpolation steps are performed to give the next 20 nodal features. This process is performed a total of 6 times, resulting in 120 features at each node after the concatenation step. These nodal features are passed as input through an MLP with 3 layers of 96, 128, and 96 hidden neurons, respectively, with ReLU activation at each hidden layer. These filter and MLP sizes were selected during the tuning process and should be selected appropriately for a given scalar field prediction problem.

The models were trained for 50 epochs with a learning rate of 0.001 using the Adam optimizer. \cite{adam} in PyTorch \cite{pytorch}. The final MLP evaluates on each node individually, but we pass in each shape during training as a mini-batch of nodes, which should induce more efficient training than pure stochastic gradient descent \cite{minibatch}. Each epoch is therefore a randomly-ordered pass through all training set meshes, inputted batch-wise into the model, one-by-one. Loss is the mean-squared error (MSE) in von Mises stress prediction across the entire shape, given by

\begin{equation}
    L(\boldsymbol \sigma, \boldsymbol f)  = \frac{1}{n}\sum_{i=1}^n \left( \sigma_i - f_i \right)^2
\end{equation}

where $\sigma_i$ is the $i$th element of ground truth field values $\boldsymbol \sigma$, $f_i$ is the $i$th element of predicted field values $\boldsymbol f$, and $n$ is the number of nodes in the mesh. Because this loss function operates across all nodes, it may be preferred over a loss function that averages across pixel/voxels. It penalizes incorrect predictions without compromising resolution of fine details, and it inherently allows areas of low node density to be considered less important to the final prediction. It should be noted, however, that varying node density throughout each part to yield more accurate solutions in critical areas is not the focus of our current work.

\begin{figure}[t]
\centering
\includegraphics[width=0.8\textwidth]{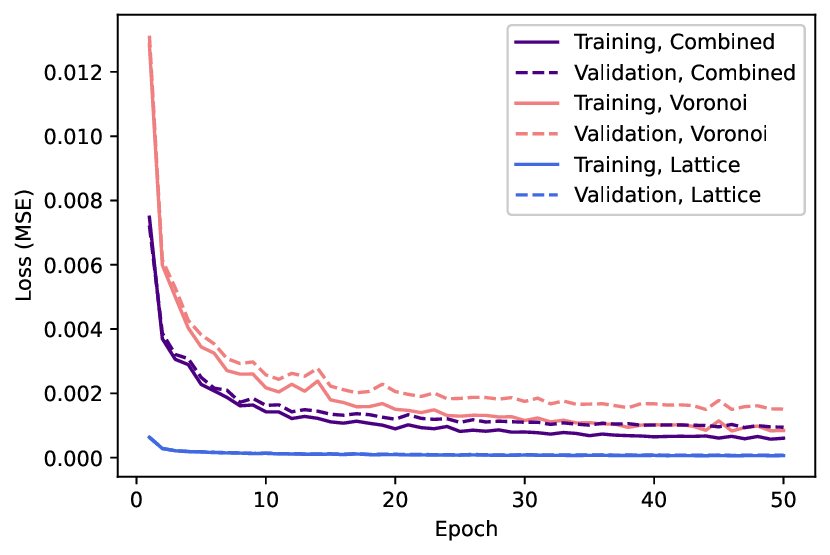}
\caption{Loss during each epoch of training for models trained on each dataset}
\label{fig:training}
\end{figure}

Figure \ref{fig:training} shows the convergence of loss during training for both datasets. Note that although there is some difference between training and validation loss, both exhibit convergence. Between datasets, however, the model trained on the Lattice Set converged to a smaller loss than the Voronoi Set model. This may indicate that our method performs better on certain types of shapes than on others. However, the typical magnitude of von Mises stress fields was lower for meshes in the Lattice Set than for the Voronoi Set. Hence, a comparison of MSE values may be misleading, so in subsequent sections we instead look at $R^2$ (defined in Sec.\ref{sec:stress}) to compare goodness-of-fit more fairly across different models/datasets.

Training a single model on the Combined Set took 68 minutes on an Intel Core i7-11700 CPU. Generating a finite element stress dataset with 2000 shapes in MATLAB took approximately 13 minutes. Using the model to predict all 2000 stress fields takes 20 seconds. Thus, once trained, using the model as a finite element method surrogate results in about 40$\times$ speedup. For a field that requires a more intensive simulation to compute, the evaluation speed of the model will remain the same; therefore, this method offers proportionally larger speedup for more complex problems.

%% file: 05-results.tex
Here we present our results for stress field predictions, performing parametric studies to investigate the effect of varying model depth and dataset size. Next, we demonstrate the model's upsampling ability, and verify that it can make predictions of another field (steady-state temperature). The results are compared to a baseline model, an interpolated U-Net.

Because some structures result in much higher peak stresses than others, in this section we compare models using the $R^2$ goodness-of-fit measure \cite{r2},

\begin{equation}
    R^2 =  1-\frac{\sum_{i=1}^n\left(\sigma_i - f_i\right)^2}{\sum_{i=1}^n\left(\sigma_i - \overline{\sigma}\right)^2},
\end{equation}

where $\sigma_i$ is the ground truth field value at node $i$, $\overline{\sigma}$ is the mean of all ground truth field values, $f_i$ is the predicted field value at node $i$, and $n$ is the number of nodes in the mesh. To supplement an $R^2$ value, we will also plot predicted-vs-actual stress for several examples. A good model will have an $R^2$ value close to 1 and a predicted-vs-actual plot that is approximately linear with a slope 1 and y-intercept 0.


\subsection{Stress Field Prediction}

\begin{figure*}
    \centering
    \includegraphics[width=0.9\textwidth]{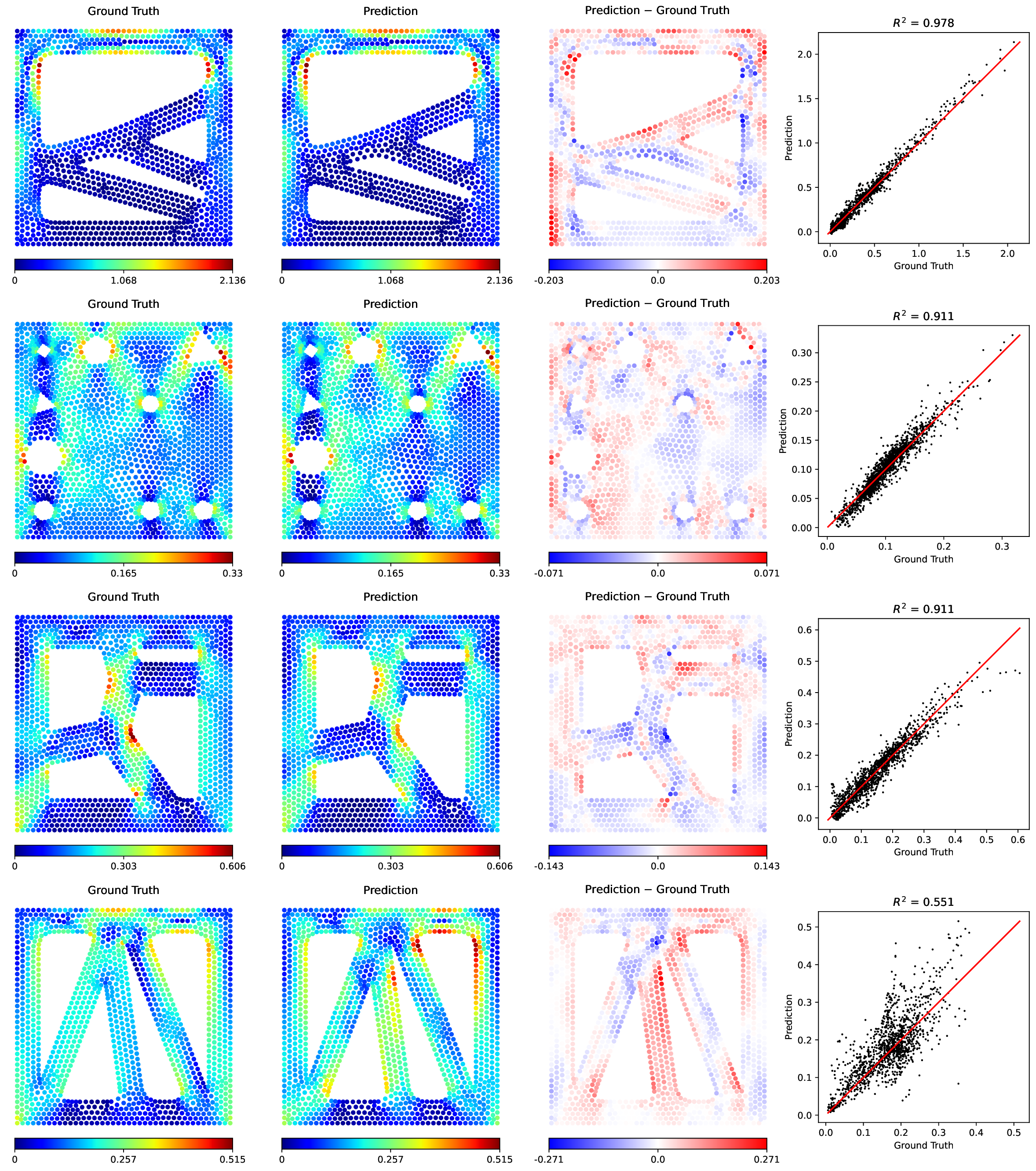}

    \caption{Stress field ground truth, prediction, difference, and predicted-vs-actual plots for testing set meshes on which the model performed (from top to bottom): 1. Best; 2. Near-median, Lattice; 3. Near-median, Voronoi; 4. Worst. Stress values are in $10^4\times$ Pascals.}
    \label{fig:visualize_stress}
\end{figure*}

Figure \ref{fig:visualize_stress} shows the ground truth field, predicted field, difference, and predicted-vs-actual plots for the best-case, median case, and the worst-case predictions in the Combined Set's testing data -- these shapes have not been seen during training, but are generated using the same parameters. 

A typical prediction captures the qualitative behavior of the stress field, accurately predicting the locations of the peak stress. For example, all four predictions shown in Fig.\ref{fig:visualize_stress} have high-stress locations that closely match their corresponding locations on the ground truth mesh, although their values may not match perfectly.

\label{sec:stress}
\begin{figure}
    \centering
    \includegraphics[width=0.8\textwidth]{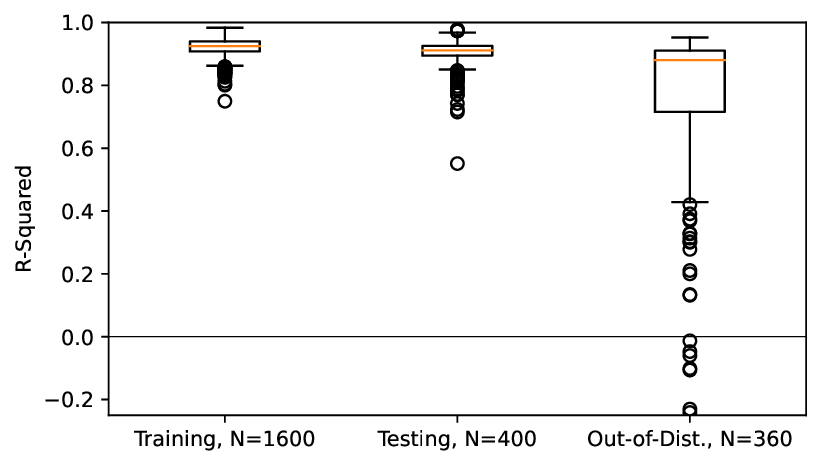}
    \caption{$R^2$ distributions for the model trained on the combined set}
    \label{fig:box}
\end{figure}

\begin{table}
\caption{Median $R^2$ for stress predictions on Voronoi, Lattice and Combined Datasets, on training, testing, and out-of-distribution data}
\centering
    \begin{tabular}{V{2} c V{2} c | c | c V{2}}\Xhline{2\arrayrulewidth}
    \multirow{2}{*}{Dataset} & \multicolumn{3}{c V{2} }{Median $R^2$}\\\cline{2-4}
        & Training & Testing & Out-of-Dist.\\\Xhline{2\arrayrulewidth}
        Voronoi Set & 0.948 & 0.921 & 0.768\\
        Lattice Set & 0.980 & 0.977 & 0.976\\
        Combined Set & 0.925 & 0.911 & 0.881\\\Xhline{2\arrayrulewidth}
    \end{tabular}
\label{tab:r2_stress}
\end{table}

 Figure \ref{fig:box} is a box-and-whisker plot showing the spread of $R^2$ values for the model trained on the Combined Set. The plot has several outliers (shown as circles), but the typical $R^2$ values are close between the training and testing sets. This indicates that a trained model can be expected to perform roughly as well on geometries in the training set as it can on a random geometry generated with the same parameters. Out-of-distribution evaluation reveals that our model may struggle to capture geometries that significantly differ from those in the training set. Further investigation into these anomalous shapes reveals that the model primarily struggles for out-of-distribution Voronoi Set geometries in which wall thickness was lower than it was in any training shapes, leading to higher stresses than the model was exposed to during training. In such cases, the model severely under-predicts stress, particularly at nodes where stress values are high.

 The median values of $R^2$ on all three partitions of both datasets are tabulated in Tab.\ref{tab:r2_stress}. A testing set median $R^2$ of 0.911 for the model trained on the Combined Set demonstrates that each model has good overall performance. Interestingly, there is not a significant decrease in prediction quality for the model trained on \textit{both} Voronoi and Lattice sets, compared to the models trained on the individual datasets. This reveals that a single model can indeed make predictions for multiple types of shapes, and it will be limited mainly by the diversity of shapes in its training dataset.


\subsection{Parametric Study: Layer Count}
For $64\times64$ input SDF images, a maximum of six pooling operations (with kernel size $2 \times 2$ and stride 2) are possible before the image is reduced to a single pixel. We refer to the number of pooling (with convolution and activation) steps as the ``layer count" of our network. We expect that increasing layer count should improve model performance. However, adding to the number of layers also increases the number of parameters, which lengthens training and evaluation time. 

Hence, we trained 6 networks on the Combined Set, with the number of pooling layers ranging from 1 to 6, to investigate this tradeoff.

\begin{table*}
\caption{Models with 1-6 pooling layers, their trainable parameter counts, training times, and median $R^2$ scores for stress predictions on the training, testing, and out-of-distribution partitions of the Combined Set}
\centering
    \begin{tabular}{V{3} c V{2} c | c V{3} c | c | c V{3}}\Xhline{3\arrayrulewidth}
    \multicolumn{3}{V{3} c V{3} }{ Model Information } & \multicolumn{3}{c V{3} }{Median $R^2$}\\\Xhline{2\arrayrulewidth}
    Pooling Layers & Parameters & CPU Training Time & Training & Testing & Out-of-Distribution\\\Xhline{2\arrayrulewidth}
        1 & 36,973 & 21.7 min. & -0.234 & -0.222 & -0.316\\
        2 & 48,913 & 33.2 min. & 0.526 & 0.529 & 0.444\\
        3 & 60,853 & 41.5 min. & 0.801 & 0.792 & 0.731\\
        4 & 72,793 & 50.8 min. & 0.896 & 0.880 & 0.840\\
        5 & 84,733 & 61.2 min. & 0.922 & 0.908 & 0.885\\
        6 & 96,673 & 68.1 min. & 0.925 & 0.911 & 0.881\\
        \Xhline{3\arrayrulewidth}
    \end{tabular}
\label{tab:param}
\end{table*}

Table \ref{tab:param} shows median $R^2$ scores for Combined Set training, testing, and out-of-distribution partitions for models with 6 different layer counts. It also lists the number of trainable parameters contained by each of these models, along with the time to train them for 50 epochs on the CPU. We observe that as layer count increases, the performance of the model on test data improves. However, each subsequent layer contributes less to the improvement of the model. For example, the $R^2$ difference between the 1- and 2-layer models was much greater than the difference between the 5- and 6-layer models. The out-of-distribution performance does begin to decrease for the 6-layer model, evidence of overfitting. Futhermore, this largest model does take more training/evaluation time compared to the smaller models. Therefore, while we show results only from the 6-layer model, careful consideration into this tradeoff should be taken when selecting a model architecture.


\subsection{Parametric Study: Dataset Size}

\begin{table}
    \caption{Median $R^2$ for stress predictions on training, testing, and out-of-distribution data in the Combined Set, for models trained with different amounts of training data}
\centering
    \begin{tabular}{V{2} c V{2} c | c | c V{2}}\Xhline{2\arrayrulewidth}
    \multirow{2}{*}{Size of} & \multicolumn{3}{c V{2} }{Median $R^2$}\\\cline{2-4}
    Training Set & Training & Testing & Out-of-Dist.\\\Xhline{2\arrayrulewidth}
        50 & 0.822 & 0.627 & 0.600\\
        100 & 0.808 & 0.713 & 0.626\\
        200 & 0.894 & 0.822 & 0.780\\
        400 & 0.910 & 0.858 & 0.808\\
        800 & 0.927 & 0.900 & 0.864\\
        1600 & 0.925 & 0.911 & 0.881\\
        \Xhline{2\arrayrulewidth}
    \end{tabular}

    \label{tab:train_size}
\end{table}

The Combined Set has 1600 training, 400 testing, and 360 out-of-distribution shape samples. It is important to study the effect of varying the size of the dataset to gain an intuition for how much the model will improve if more training data is provided. Therefore, we trained 6 more stress prediction models. Each of these models was provided a random subset of the in-distribution data for training, with the rest to be used for testing. The out-of-distribution data is once again not used during training, but is instead used to evaluate each trained model.

Table \ref{tab:train_size} shows the results of each model, with between 50 and 1600 shapes used for training. In every case, additional data improved the model performance on testing and out-of-distribution data. It appears as though including more shapes in training data leads to a model with better generalizing abilities. These effects are less significant for large training sets. For example, increasing the dataset size from 800 to 1600 shapes only yielded an increase in median test data $R^2$ from 0.900 to 0.911. Such a marginal improvement may not be worth it if it means generating significantly more training data.


\subsection{Upsampling to a Higher Mesh Resolution}

\begin{figure*}
    \centering
    \includegraphics[width=0.9\textwidth]{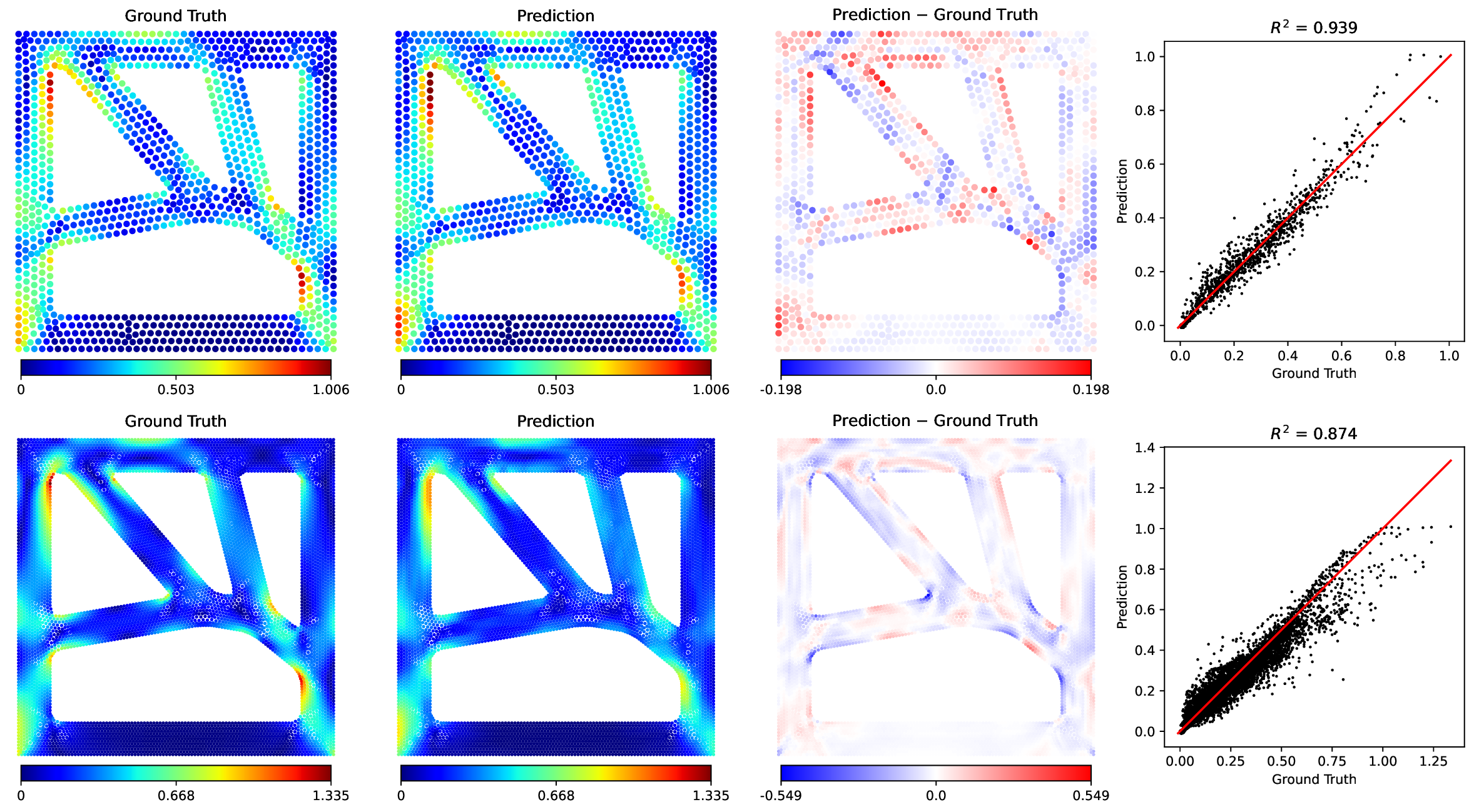}
    
    \caption{Ground truth, prediction, difference, and predicted-vs-actual plots for coarse (top) and fine (bottom) meshes. Stress values are in $10^4\times$ Pascals.}
    \label{fig:coarse_fine}
\end{figure*}

We claim that the mesh-independence of our method enables a model trained on a coarse-resolution mesh to make a prediction on a fine-resolution mesh without any additional training. After all, the model performs separate predictions for each node. On the original datasets, the maximum edge length on each mesh was set to 0.025m. Generating a mesh on the Voronoi Set with a maximum edge length of 0.01m (roughly 6 times as many nodes per mesh), we can make a stress field prediction on the finer mesh because interpolation is built into the network. Figure \ref{fig:coarse_fine} shows a visualization of a coarse and fine mesh for the same Voronoi geometry, evaluated by the model trained on the Combined Set. Note that the field generated is similar for both, demonstrating the implicit interpolation capabilities of our method.

It should be noted that the fidelity of the ML result is limited to the resolution of meshes in the training data, and will not match high-resolution finite element results; instead, this analysis demonstrates the flexibility that the method offers. Once trained for a physical problem, an engineer can use any mesh generation technique and any mesh resolution, and the model will make a prediction according to the physical problem on which it was trained.


\subsection{Alternate Scalar Field: Predicting Temperature}

\begin{table}
\caption{Median $R^2$ for temperature predictions on Voronoi, Lattice and Combined Datasets, on training, testing, and out-of-distribution data}
\centering
    \begin{tabular}{V{2} c V{2} c | c | c V{2}}\Xhline{2\arrayrulewidth}
    \multirow{2}{*}{Dataset} & \multicolumn{3}{c V{2} }{Median $R^2$}\\\cline{2-4}
        & Training & Testing & Out-of-Dist.\\\Xhline{2\arrayrulewidth}
        Voronoi Set & 0.995 & 0.992 & 0.951\\
        Lattice Set & 0.995 & 0.994 & 0.993\\
        Combined Set & 0.990 & 0.988 & 0.976\\\Xhline{2\arrayrulewidth}
    \end{tabular}

\label{tab:r2_temp}
\end{table}

\begin{figure*}
    \centering
    \includegraphics[width=0.9\textwidth]{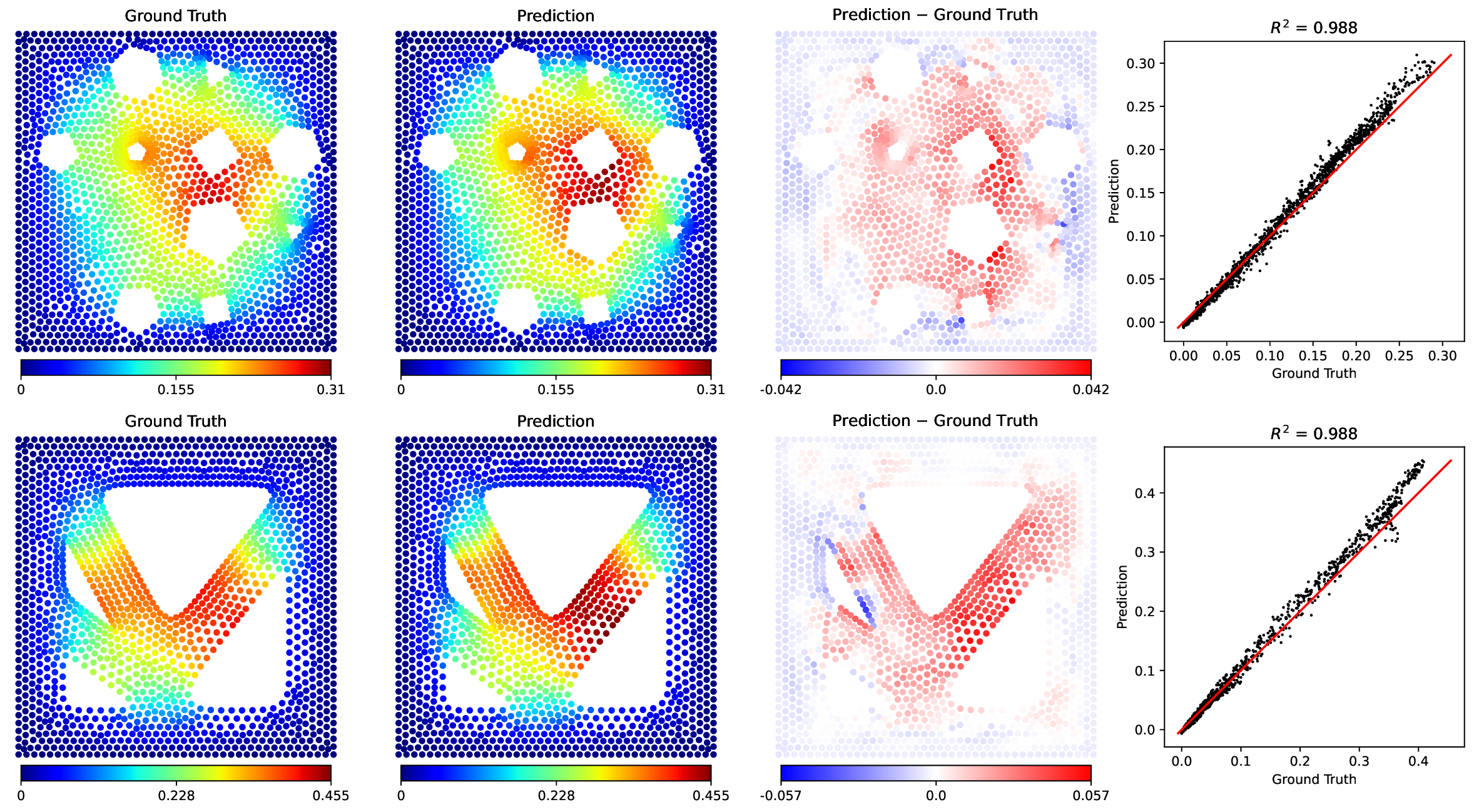}
    \caption{Temperature field ground truth, prediction, difference, and predicted-vs-actual plots for testing set meshes with near-median performance. Temperature values are in $^{\circ}$C.}
    \label{fig:visualize_temperature}
\end{figure*}

We have made the claim that by demonstrating our method on a von Mises stress field, we can expect similar results for other fields present in engineering problems. As an example of another field of interest, we will also show that our method is effective at predicting a temperature field in a steady-state heat transfer setting. 

We therefore pose a separate problem on the same 2-D domain: The part has the thermal conductivity of aluminum, and all four outer boundary walls have a fixed temperature constraint of 0 $^{\circ}$C. Each pore now acts as a heat source with constant heat flux 100 W/m$^2$. We assume unit part thickness. These quantities are once again chosen arbitrarily, and although they result in very low temperature values across the mesh -- from 0 $^{\circ}$C to about 0.6 $^{\circ}$C -- the distribution has enough variation that we can still judge whether our method is capable of predicting a temperature field.

\begin{table*}[htbp]
\caption{Median $R^2$ for (baseline) Interpolated U-Net compared to (our) Interpolated Multi-Resolution CNN on each partition of the Combined Set for the stress and temperature problems}
\centering
    \begin{tabular}{V{3} c V{3} c V{3} c | c | c V{3}}\Xhline{3\arrayrulewidth}
    \multirow{2}{*}{Field} & \multirow{2}{*}{Model} & \multicolumn{3}{c V{3} }{Median $R^2$}\\\cline{3-5}
        & & Training & Testing & Out-of-Distribution\\\Xhline{3\arrayrulewidth}
        \multirow{2}{*}{Stress} & Interpolated U-Net & 0.834 & 0.833 & 0.809\\
        & \textit{Interpolated Multi-Resolution CNN} & 0.925 & 0.911 & 0.881\\\hline
        \multirow{2}{*}{Temperature} & Interpolated U-Net & 0.895 & 0.897 & 0.861\\
        & \textit{Interpolated Multi-Resolution CNN} & 0.990 & 0.988 & 0.976\\\Xhline{3\arrayrulewidth}
    \end{tabular}

\label{tab:unet}
\end{table*}

Table \ref{tab:r2_temp} contains the temperature prediction $R^2$ values for all three partitions of models trained on the Voronoi, Lattice, and Combined Sets. The temperature fields are noticeably smoother than the von Mises stress fields, exhibiting a lack of steep spatial gradients, which has a noticeable effect on prediction quality. This can be observed in Fig.\ref{fig:visualize_temperature}, which depicts two typical predictions in the testing set for the ``combined" model. This observation is in accordance with the better performance of our method, with testing set median $R^2$ values of 0.992 and 0.994 for the individual datasets' models and 0.988 for a model trained on both.

For nodes with larger temperature values, the model appears to slightly over-predict the temperature, an effect visible in both the field visualizations and predicted-vs-actual plots. Effects like this can likely be attributed to the specifications of the problem being solved. For example, for this temperature problem, the edges of the square have a fixed low temperature across different shapes, so making a prediction near an edge is less challenging. However, the holes in the center of each shape introduce a heat flux, causing changes in temperature that are less predictable than a fixed temperature condition. Thus, care should be taken to ensure that problem-specific artifacts do not have too much influence on the predictive model.


\subsection{Comparison with Interpolated U-Net}

A standard fully-convolutional U-Net \cite{unet} whose final output is interpolated to each node can serve as a baseline method for benchmarking the performance of our multi-resolution model. Therefore, we train U-Net models for the stress and temperature problems on the Combined Set and compare the median $R^2$ results with those of the our proposed model.

The U-Net takes the same $64\times64$ SDF image as input. It undergoes two $3\times3$ convolutions (each followed by ReLU activation and batch normalization) and the resulting feature maps are pooled using max-pooling. There are 5 such convolution/pooling layers, with channel sizes 12, 18, 24, 30, and 36 respectively, which make up the encoder side of the network. Then a set of max-unpool and convolution steps form the decoder, with concatenation skip-connections to preserve fine details across the network. A final convolution reduces the channel count to 1, to predict a scalar field, which must be interpolated to node coordinates. Here, we use the same bilinear interpolation method as in our proposed model. The network size parameters have been chosen to give a model with size comparable to our 6-layer, 96,673-parameter interpolated multi-resolution CNN; the U-Net has 5 convolution/pooling layers and 99,217 trainable parameters.

Table \ref{tab:unet} reveals that the Interpolated Multi-Resolution CNN is an improvement over a similarly-sized Interpolated U-Net for both the stress and temperature prediction problems. It had slightly better $R^2$ values for training, testing, and out-of-distribution dataset partitions. Therefore, we believe interpolating the features separately at each resolution is more effective than using a single interpolation at the final step. In addition, the inclusion of an MLP at the end of the network, as opposed to the purely convolution-based U-Net, may yield a more powerful model. One drawback of our method is that model training/evaluation is slower than an interpolated U-Net by approximately $1.5\times$, which can be attributed to the additional interpolation steps and the MLP. Therefore, the improvement in accuracy should be weighed against the added runtime when selecting a scalar field prediction model.

%% file: 06-limitations.tex
One drawback of our proposed method is that each trained model can make predictions of one type of field for a single material in one loading case. Future work will look at parametric variation of loads and boundary conditions. Material properties, including those for nonhomogeneous materials, can also be inputted as features. Prediction of other fields, including vector fields or even time-varying fields, may be investigated as well.

Although the model has been defined and demonstrated in a 2-D context, 3-D geometries are more common throughout engineering. Both the convolution and the interpolation steps in the model can be extended to 3-D, and the SDF will still serve as a viable input in higher-dimensional settings. Therefore, we are interested in adapting the model to make predictions for 3-D meshes and testing it on a suitable 3-D dataset.

Restriction of the geometry to a $1\text{m}\times 1\text{m}$ square envelope is also a significant geometric constraint, but this type of constraint is reasonable in that it mimics many real design decisions in engineering. In spite of this, we are still interested in expanding our method to operate on shapes with fewer restrictions.

%% file: 07-conclusion.tex
In this work, we proposed a model that makes predictions for scalar fields by interpolating the intermediate results of a multi-resolution CNN at nodal locations, and then passing the resulting feature vectors through an MLP. The model input is a signed distance field sampled on a fixed grid, while the output is a scalar field prediction at arbitrary nodal locations.

Two 2-D datasets were created, the Voronoi Set and the Lattice Set, each of which contain shapes that were designed to be difficult to represent using a small number of parameters. Next, our model was trained on a combined dataset of these shapes to predict the von Mises stress field for a part undergoing compression. Our model achieved an overall median $R^2$ of approximately 0.91 for stress field prediction on the testing set. Training on a steady-state temperature problem, the model has even stronger performance, yielding $R^2$ values above 0.98. Thus, the proposed method demonstrates the potential to replace finite element computations with less expensive surrogate model evaluations in an engineering design setting. For the linear elasticity problem we demonstrated, speedup by using the model instead of performing a finite element solve is roughly $40\times$.

For $64\times 64$ SDF input, a network with 6 pooling steps was shown to give the best performance, at the expense of slightly longer training/evaluation time compared to smaller models. This model gives better results than the baseline model, an interpolated U-Net. Because the network contains interpolation steps, it can generate output for any collection of nodes, even those of very fine meshes.

Future work will explore variation of boundary conditions and expansion to 3-D geometries, for a more robust field prediction model.